\documentclass[cameraready]{Interspeech}

\title{Fast When, Careful Who: Dual-Process Multiparty Turn-Taking with Diffusion Augmentation}

\author[affiliation={1}, orcid=0000-0002-6397-0974]{Rutherford A.}{Patamia}
\author[affiliation={1}, orcid=0000-0002-2160-6111, correspondingauthor]{Ming}{Liu}
\author[affiliation={1}, orcid=0000-0002-4711-7543]{Wei}{Luo}
\author[affiliation={2}, orcid=0000-0003-4986-4953]{Favour}{Ekong}
\author[affiliation={1}, orcid=0000-0003-4203-6477]{Akan}{Cosgun}

\address{
    $^1$ Deakin University, Melbourne, Australia \\
    $^2$ Griffith University, Brisbane, Australia
}

\email{r.patamia@deakin.edu.au}


\usepackage{comment}
\usepackage{amsmath}
\usepackage{multirow}

\begin{document}
\maketitle

\begin{abstract}
Reliable turn-taking is essential for spoken dialogue systems. However, most existing methods are designed for two-speaker interaction and struggle with realistic multiparty audio containing overlap and rapid speaker changes. We study multiparty turn-taking on the VoxConverse dataset and propose an audio-only two-stage pipeline that separates when to trigger a turn boundary from whether the floor is actually transferring. A fast trigger scans the audio and proposes candidate end-of-turn times, while a lightweight verifier runs only at those times to decide \textsc{Hold} or \textsc{Shift} and support next-speaker prediction. We report results in the full multiparty setting and a controlled dyadic top-2 projection for comparability. We also investigate diffusion-based, label-preserving background-audio mixing as a data augmentation strategy. Results show improved shift detection over a baseline, with further improvements from diffusion augmentation.
\end{abstract}

\section{Introduction}
Spoken dialogue systems are increasingly expected to support full-duplex interaction, where users may overlap, backchannel, and yield the floor without explicit push-to-talk cues. In multiparty environments, such functionality presents a critical control challenge: a system must decide \emph{when} to respond (hold vs. transfer) and, when transfer is likely, \emph{who} will speak next. Errors in turn timing can disrupt deployment; premature entries cause overlaps, while delays result in missed handovers. Evaluation should consider streaming constraints instead of treating turn boundaries solely as labels for offline scoring, which assesses predictions after the full recording is available \cite{fdb}.

A recent approach to turn-taking prediction, Voice Activity Projection (VAP) \cite{vap}, forecasts near-future voice activity from ongoing audio to infer events like pauses and overlaps. This self-supervised task predicts a discretised future voice-activity window and evaluates turn-taking through zero-shot tasks, mainly focusing on dyadic interactions. Subsequent improvements allowed real-time, frame-synchronous VAP inference and examined the accuracy-latency trade-off in shorter CPU contexts \cite{rva}. More recently, projection-style modelling has been adapted to triadic interactions, but it struggles to scale due to the increase in the number of discrete joint states \cite{tva}. Despite these advancements, projection methods primarily infer transitions from activity forecasts and are often evaluated on small, fixed sets of speakers. In realistic multiparty settings, like the VoxConverse dataset \cite{vox}, which averages 4.5 speakers in development and 6.5 in testing, short pauses and overlaps create challenging activity patterns that require additional speaker-specific evidence at decision points.

This paper adopts a fast-slow dual-process framework for time-sensitive interactions \cite{dua}, emphasising the tension between responsiveness and capability. Fast mechanisms react quickly but can be brittle, while deliberative ones, though accurate, are resource-intensive. In turn-taking scenarios, continuously using highly discriminative reasoning is computationally demanding, and relying solely on rapid local cues can lead to premature turn grabs. Unlike end-to-end models that must continuously learn when to trigger and who is next from limited multiparty supervision, the dual-process approach separates candidate decision times (potential turn boundaries) from candidate verification, determining if the current speaker continues (HOLD) or transfers the floor (SHIFT), and predicting the next speaker upon confirmation.

Specifically, the fast framework serves as an end-of-turn \emph{gate}, a low-latency detector that scans the audio stream for candidate decision points using WavLM \cite{wlm} with a minimally trainable head. The slow framework or verifier only engages at these candidate times to compute short-context speaker evidence with a speaker-embedding model \cite{eca}, testing for speaker change and ranking likely next speakers. The gate controls the decision frequency, while the verifier selectively determines whether candidates are accepted.

Our primary contribution is the study of label-preserving diffusion augmentation through background-audio mixing, which incorporates diffusion-generated background audio into training waveforms while maintaining candidate times and turn labels \cite{dif,diff,ges}. Our secondary contribution is an audio-only two-stage pipeline for multiparty turn-taking on the VoxConverse dataset \cite{vox}, separating (i) low-latency boundary triggering from (ii) speaker-aware floor-transfer verification and next-speaker ranking, with results reported under the full multiparty condition. Unlike prior diffusion-based augmentation that synthesises additional examples for downstream tasks, our method perturbs acoustics without altering the turn structure \cite{dif1,dif2,dif3}. For a controlled benchmark comparison, we evaluated a dyadic projection baseline (VAP) using the same candidate-time protocol, tuning hyperparameters on validation data and reporting results on the held-out test set \cite{vap,rva,tva}.

\section{Method}
\subsection{Dataset}
We evaluate on VoxConverse~\cite{vox}, a YouTube-derived multiparty diarisation corpus notable for natural overlap and diverse acoustics. This dataset was selected for its public availability, manageable size, and inclusion of debate-style segments with rapid speaker exchanges. While VoxConverse is audio-visual, we focus solely on the audio and ground-truth Rich Transcription Time Marked (RTTM) annotations, which include timestamped speech segments with speaker IDs but no transcriptions. The corpus consists of 216 development recordings (1,218 minutes; $\sim$4.5 speakers/rec) with approximately 3–4\% overlapped speech time~\cite{vox}. Candidate decision points are marked at RTTM segment boundaries, classified as \textsc{HOLD} if the next segment is by the same speaker and \textsc{SHIFT} otherwise.

Our main interest is a full multiparty experiment. For a controlled dyadic comparison with Voice Activity Projection, we select the two speakers with the most speech time and evaluate our model against Voice Activity Projection on the same timestamps, using recording-level train/validation/test splits to prevent data leakage.

\subsection{Candidate Construction and Preprocessing}
We generate labels for the complete multiparty setting using ground-truth RTTM diarization, since VoxConverse indicates who spoke when but lacks specific turn-taking labels such as SHIFT/HOLD. We refine each speaker’s RTTM segments into clearer “turns”; if two segments from the same speaker are separated by a brief silence, they are treated as one continuous turn. For example, segments [10.00–10.40] and [10.43–10.90] from speaker A are combined into a single turn spanning [10.00–10.90]. Each turn's end is marked as a decision moment, labelled SHIFT if another speaker takes the floor, or HOLD if the same speaker resumes. For training and controlled comparison with Voice Activity Projection \cite{vap}, these decision moments and labels remain consistent across methods during evaluation.

In the dyadic top-2 benchmark, we do not merge speakers or create new transitions; we retain only decision moments involving the two most talkative participants and disregard others. Diffusion background mixing is applied to the training audio post-definition of decision moments and labels, ensuring that times and labels remain unchanged.

\subsection{Dual-Process Turn-Taking Framework}\label{sec:framework}
\begin{figure}[t]
  \centering
  \includegraphics[width=\linewidth]{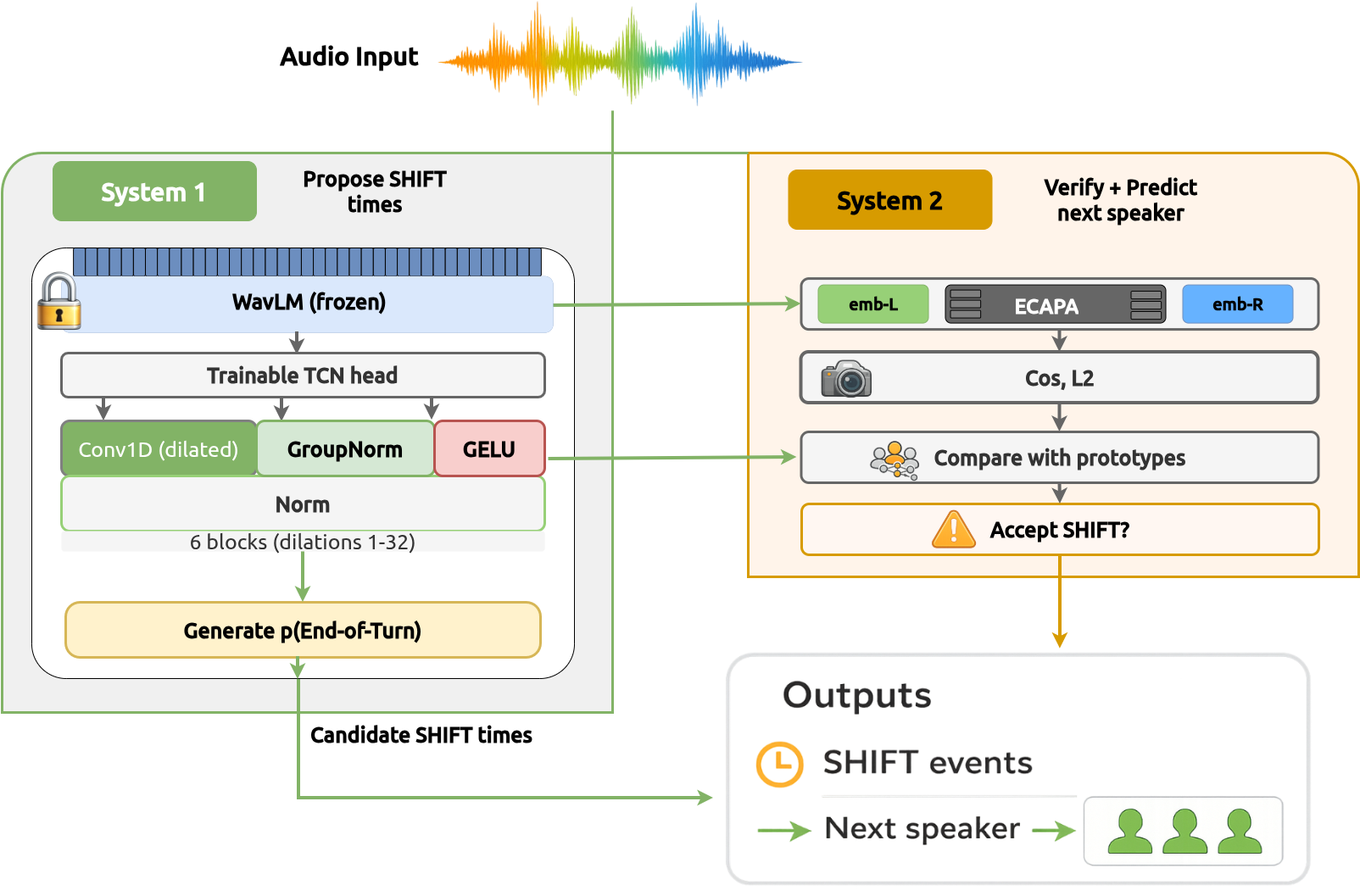}
  \caption{Dual-process model architecture}
  \label{fig:ma}
\end{figure}

Dual Process Theory (DPT) distinguishes between two types of cognitive processing \cite{dua2}: System~1 processes are fast and automatic, producing default responses with minimal working-memory demand, while System~2 processes are deliberate and rely on working memory for hypothetical reasoning. Recent applications of DPT, such as DPT-Agent, utilise a fast controller (a finite-state machine) for low-latency actions, while a slower module employs a large language model (LLM) for Theory-of-Mind inference and strategy revision. This system's effectiveness is assessed in real-time human-AI collaboration tasks \cite{lev}. DPT is manifested as a proposal–verification pipeline: System~1 proposes end-of-turn times, and System~2 verifies floor transfer and who-next at those candidate times. As depicted in Figure~\ref{fig:ma}

\textbf{System 1 (proposer): end-of-turn gate.}
Let $x(t)$ be the mixed audio waveform. System~1 scans the audio stream using a sliding window of duration $\Delta$ and step size (``hop'') $h$ seconds, and outputs an updated end-of-turn \textit{score} at regular intervals of \(h\) seconds. Specifically, with \( \Delta = 1.0 \) s and \( h = 0.1 \) s, the system computes a new score based on the most recent 1.0 s of audio data. This configuration results in significant overlap across consecutive hops (yielding 10 updates per second). The end-of-turn \textit{score} at time \( t \) is defined mathematically as follows:
\begin{equation}
p_{\mathrm{eot}}(t)=g_{\theta}\bigl(\phi(x_{t-\Delta:t})\bigr).
\end{equation}
Here, $\phi(\cdot)$ is a pretrained WavLM encoder that maps a waveform window to speech embeddings, and $g_\theta(\cdot)$ is a small trainable head that maps those embeddings to $p_{\mathrm{eot}}(t)\in[0,1]$, interpreted as an end-of-turn likelihood score.
To translate the frame-level score stream into discrete decision points, we apply a straightforward persistence-and-refractory rule. A candidate time $\tau$ will be emitted only if the score remains above a predefined threshold $\eta$ for $K$ consecutive hops (temporal persistence) and if $\tau$ occurs at least $\Delta$ seconds after the preceding candidate (cooldown/refractory period). This can be expressed as:
\begin{equation}
\tau\in\mathcal{C}\iff \bigwedge_{j=0}^{K-1} p_{\mathrm{eot}}(\tau-jh)\ge\eta\ \land\ (\tau-\tau_{\mathrm{prev}})\ge\delta.
\end{equation}
In our implementation, audio is processed with $\Delta=1.0$s windows and a hop size $h=0.1$s; at test time, we use $K=3$ and $\delta=0.8$s (so the score must stay above the threshold for about $3\times 0.1=0.3$s before we trigger a candidate, and we then wait at least $0.8$s before allowing another trigger).

\textbf{System 2 (verifier): speaker-change check (and who-next).}
For each proposed candidate $\tau\in\mathcal{C}$,  System~2 extracts short waveform contexts immediately \textit{before} and \textit{after} $\tau$ and compares speaker evidence, i.e., whether the same speaker identity is present on both sides of the candidate. Specifically, a pretrained speaker-embedding extractor, namely ECAPA-TDNN \cite{eca}, denoted as $\zeta(\cdot)$, which maps each waveform segment to a fixed-dimensional vector that is discriminative to speaker identity. We then define a same-speaker score by comparing the pre/post embeddings:
\begin{equation}
p_{\mathrm{same}}(\tau)=v_{\psi}\bigl(\zeta(x_{\tau-T:\tau}),\,\zeta(x_{\tau:\tau+T})\bigr).
\end{equation}
where $T$ represents the half-width of the pre/post context (we use $T{=}1.5$s). In our default instantiation, $v_\psi$ is a lightweight similarity function computed on top of the ECAPA-TDNN embeddings (ECAPA-TDNN itself only produces embeddings), we compute cosine similarity between the pre- and post-$\tau$ embeddings and rescale it as a bounded same-speaker score $p_{\mathrm{same}}(\tau)\in[0,1]$,  where higher values indicate stronger evidence that the same speaker is continuing. This score is then subjected to a threshold $\gamma\in[0,1]$ (optimized on the validation set), we predict \textsc{SHIFT} if $p_{\mathrm{same}}(\tau)<\gamma$ and \textsc{HOLD} otherwise, with $\gamma$ fine-tuned on validation data. 
 
 Importantly, the same ECAPA embeddings are also utilised for predicting the next speaker. In multiparty scenarios, we assume an oracle speaker inventory and construct per-recording ECAPA prototypes by averaging embeddings from multiple enrollment segments for each speaker (sourced from RTTM speech). Following a validated  \textsc{SHIFT}, we embed a short post-$\tau$ context and score it against all speaker prototypes (e.g., using cosine similarity and/or $\ell_2$ distance). The most similar prototype is then identified as the predicted next speaker.
 
\subsection{Model Components and Training}
System~1 uses a fixed, pretrained speech representation model and a trainable temporal head to generate end-of-turn scores at each time step. The head is trained with binary cross-entropy and the AdamW optimiser over 20 epochs, with checkpoints based on validation AUPRC and optimised decision thresholds described in section~\ref{sec:framework}. System~2, operating at proposed decision times, employs a pretrained ECAPA-TDNN speaker-embedding model to extract embeddings from 1.5 seconds before and after the decision point. Cosine similarity between these embeddings indicates speaker continuity (HOLD) or change (SHIFT), with thresholds optimised for SHIFT performance. For next-speaker prediction, prototype embeddings are averaged from RTTM segments, and after a confirmed SHIFT, the highest similarity embedding is selected. To enhance acoustic diversity, diffusion augmentation stochastically mixes a diffusion-generated background audio track into the training waveforms with a 0.6 probability, using an SNR sampled between 10–25 dB, while ensuring that timestamps and HOLD/SHIFT labels remain consistent. 

For comparison, we also train a lightweight two-stage baseline that mirrors the pipeline structure but replaces pretrained backbones with (i) a residual CNN proposer and (ii) a Siamese-style verifier \cite{sia}. Residual CNNs serve as compact baselines for audio modelling, and Siamese encoders are well-suited for similarity-based speaker verification. We train and tune the baseline using the same train/validation/test splits and validation-only threshold selection, reporting results both without and with diffusion augmentation to assess background mixing effects.

\section{Experimental Results}
We evaluate on VoxConverse development mixture audio using Oracle RTTM annotations and a fixed recording-level split used by our candidate-time protocol (194 sessions total: \%70/20/10 train/validation/test). Although the development partition contains more recordings, our split lists are built from the subset that passes the preprocessing required by our evaluation protocol; sessions are omitted when the pipeline cannot generate the necessary oracle decision-time annotations for that recording (e.g., missing audio/RTTM files or other preprocessing failures). We keep this fixed subset to ensure a consistent, reproducible evaluation across all methods and comparisons. 

\subsection{Metric}\label{subsec:metr}
We evaluate two aspects: (i) detection and timing of SHIFT events, as these are essential for verifying turn transfers in our pipeline, and (ii) prediction of the next speaker. Our SHIFT results are reported against the complete multiparty ground truth (GT\_any), which includes all speaker transitions. For SHIFT detection, we match predicted shift times to ground-truth times within a tolerance window. Each prediction and ground-truth event can be used once; if multiple predictions fall within the same window, the closest one is matched, and the rest are counted as false positives. A true positive corresponds to a ground-truth shift, while an unmatched prediction is a false positive, and an unmatched ground-truth shift is a false negative. We report precision, recall, F1 scores, and mean absolute timing error (MAE) for matched shifts.

Next-speaker predictions are assessed in two modes: End-to-End (E2E on matched shifts) evaluates who speaks next only on correctly matched shifts, reporting accuracy, macro-F1, and coverage (the proportion of ground-truth shifts matched). Ground-Truth (GT)-boundary (oracle-timed) evaluates next-speaker prediction at actual ground-truth shift boundaries, separating speaker identification from boundary detection.

For the controlled dyadic VAP \cite{vap} benchmark, we also report balanced accuracy (BAcc), the average of the true positive rate and true negative rate, as this metric is less sensitive to class imbalances in dyadic HOLD/SHIFT decisions.

\subsection{Shift detection}
Table~\ref{tab:csns} indicates that the lightweight baseline often over-triggers in multiparty audio, resulting in numerous false alarms during brief pauses, backchannels, and overlapping speech. Our two-stage pipeline mitigates these erroneous triggers by verifying proposed turn transfers against speaker evidence, resulting in a more selective boundary stream while maintaining sensitivity to genuine floor transfers. The incorporation of diffusion-based background mixing further enhances robustness in noisy and overlapping segments and reduces timing errors for matched shifts. The remaining errors are primarily found in heavily overlapped, fast-exchange areas, where end-of-turn cues are acoustically ambiguous, and multiple speakers vie for the floor.  

To conduct a controlled dyadic benchmark comparison (see Table~\ref{tab:a_2_a}), we set up an environment that removes any advantages that might come from varying the number of decision times. This approach allows us to isolate the effectiveness of the HOLD/SHIFT decision at consistent intervals. In both methods (dual-process design and voice activity projection), we implement validation-only tuning: model parameters are fixed, and we optimise the decision threshold on the validation split. We apply the same-speaker threshold for our verifier and the shift/hold score threshold for VAP before presenting the results on the test split.

\begin{table}[!htbp]
\centering
\caption{Multiparty SHIFT Detection results with/without diffusion background mixing}
\label{tab:csns}
\scriptsize
\setlength{\tabcolsep}{3.2pt}
\renewcommand{\arraystretch}{1.08}

\begin{tabular}{p{2.05cm}cccc}
\hline
\multicolumn{5}{c}{\textbf{SHIFT Detection}} \\
\hline
\textbf{Method} & \textbf{PRE} & \textbf{REC} & \textbf{F1} & \textbf{MAE} \textbf{$\pm$} \textbf{std} \\
\hline
Baseline & 0.211 & 0.260 & 0.233 & 0.194$\pm$0.158  \\
Baseline (+DiffAug) & 0.342 & 0.338 & 0.340 & 0.186$\pm$0.146 \\
Ours & 0.567 & 0.494 & 0.528 & 0.189$\pm$0.134 \\
Ours (+DiffAug) & \textbf{0.714} & \textbf{0.571} & \textbf{0.635} & \textbf{0.131$\pm$0.118} \\
\hline
\end{tabular}
\vspace{0.75em}
\end{table}

\begin{table}[!htbp]
\centering
\caption{Two-speaker SHIFT detection results}
\label{tab:a_2_a}
\scriptsize
\setlength{\tabcolsep}{3.2pt}
\renewcommand{\arraystretch}{1.08}

\begin{tabular}{p{2.05cm}cccc}
\hline
\textbf{Method} & \textbf{PRE} & \textbf{REC} & \textbf{F1} & \textbf{BAcc} \\
\hline
VAP \cite{vap} & 0.452 & 0.447 & 0.449 & 0.609 \\
Ours & \textbf{0.539} & \textbf{0.734} & \textbf{0.622} & \textbf{0.735} \\
\hline
\end{tabular}
\end{table}

\subsection{Next-speaker prediction}
Table~\ref{tab:nextspeaker} presents the next-speaker performance across the two evaluation modes outlined in Section~\ref{subsec:metr}. In the GT-boundary (oracle-timed) mode, our speaker selection approach demonstrates strong performance, indicating that, when the shift time is predetermined, identifying the next speaker is relatively consistent. In the End-to-End (E2E on matched shifts) mode, the evaluation of the "who-next" metric is restricted to correctly detected shifts, meaning that overall performance is predominantly constrained by shift coverage rather than speaker attribution: missed shifts limit the set of events available for scoring who-next. Additionally, diffusion augmentation appears to have minimal impact on oracle-timed who-next performance, suggesting that the advantages seen in the end-to-end approach largely stem from enhanced shift detection and coverage rather than improvements in next-speaker scoring.\\

\begin{table}[t]
\centering
\scriptsize
\setlength{\tabcolsep}{3.0pt}
\renewcommand{\arraystretch}{1.10}

\caption{Next-speaker in two modes: E2E (matched shifts; Cov.) and GT-boundary (oracle shift times); baseline is ResCNN–Siamese.}
\label{tab:nextspeaker}

\begin{tabular}{lccc cc}
\hline
\multicolumn{6}{c}{\textbf{NEXT-SPEAKER (E2E on matched shifts)}} \\
\hline
\multirow{2}{*}{\textbf{Method}} &
\multirow{2}{*}{\textbf{Cov.}} &
\multirow{2}{*}{\textbf{Acc$_\mu$}} &
\multirow{2}{*}{\textbf{MF1$_\mu$}} &
\multicolumn{2}{c}{\textbf{Session (mean$\pm$std)}} \\
\cline{5-6}
 &  &  &  & \textbf{Acc} & \textbf{MF1} \\
\hline
Baseline  & 0.238 & 0.352 & 0.268 & 0.316$\pm$0.281 & 0.316$\pm$0.584 \\
Baseline (+DiffAug) & 0.401 & 0.444 & 0.525 & 0.472$\pm$0.450 & 0.333$\pm$0.447 \\
Ours     & 0.533 & 0.976 & 0.976 & 0.908$\pm$0.244 & 0.894$\pm$0.266 \\
Ours (+DiffAug)  & \textbf{0.554} & \textbf{0.992} & \textbf{0.994} & 0.947$\pm$0.223 & \textbf{0.947$\pm$0.223} \\
\hline
\end{tabular}

\vspace{0.55em}

\begin{tabular}{lccc}
\hline
\multicolumn{4}{c}{\textbf{NEXT-SPEAKER (GT-boundary / oracle-timed)}} \\
\hline
Method & Cov. & Acc$_\mu$ & MF1$_\mu$ \\
\hline
Baseline        & 1.000 & 0.630 & 0.645 \\
Baseline (+DiffAug) & 1.000 & 0.505 & 0.545 \\
Ours   & 1.000 &\textbf{ 0.918} & 0.892 \\
Ours (+DiffAug) & 1.000 & 0.915 & 0.892 \\
\hline
\end{tabular}

\end{table}

\section{Discussion}
In this work, we discovered that the performance of our end-to-end pipeline, specifically, the evaluation of the next speaker based only on shifts that our system detects and matches, is mainly limited by boundary coverage, the proportion of actual shifts that our system successfully detects and matches for evaluation purposes. When boundary timing is provided (oracle-timed; the ground-truth shift timestamp), our next-speaker module performs effectively and shows little variation with diffusion augmentation. This pattern is reflected in Table~\ref{tab:nextspeaker}: oracle-timed who-next remains high, while the pipeline-mode evaluation is limited mainly by coverage. However, in an end-to-end evaluation, the accuracy of predicting who speaks next is contingent on detected and matched shifts. Consequently, any missed shifts lead to a decrease in coverage. Therefore, improving candidate quality in overlap-heavy regions is likely to yield substantial gains in end-to-end "who-next" performance.

We identified a common source of errors that aligns with established ambiguities in conversational audio. Short within-turn pauses and backchannels can easily be mistaken for the completion of a turn, while overlaps can obscure who is currently holding the floor. Our findings suggest that allocating speaker-specific computation only at likely decision points serves as a pragmatic solution in this context. This approach not only addresses the observed errors but also complements activity-projection techniques, such as VAP, which conceptualise future interactions using discretised joint activity states. Notably, as group size increases, the number of possible joint states expands exponentially, prompting the need for coarser representations in triadic settings. While recent work extends Voice Activity Projection to triadic multi-party dialogue and reports improvements over baselines on triadic conversational data, it is not directly evaluated here due to differences in data domain and the additional engineering required to adapt the method to VoxConverse’s variable-speaker multi-party setting. Rather than relying solely on projected activity patterns, our methodology leverages speaker evidence at proposed boundaries to resolve ambiguous cases. Consistent with this, Table~\ref{tab:csns} shows our pipeline operates at a more conservative boundary point than the baseline (higher precision than recall), which reduces false alarms but can miss some true handovers. This is significant for improving the accuracy of next-speaker predictions, given that our controlled dyadic comparisons against VAP are limited to decision-making at identical oracle-candidate times. 

Furthermore, the consistent mixing of diffusion backgrounds significantly enhances the boundary detection stage. This improvement leads to better SHIFT performance, reduced timing errors, and greater consistency in candidate times and labels. Table~\ref{tab:csns} directly supports this claim, indicating that diffusion enhances both SHIFT detection and timing for our system, as well as for the baseline. In contrast, oracle-timed next-speaker accuracy for our system remains largely unchanged with diffusion augmentation. This suggests that the benefits of end-to-end who-next predictions are primarily driven by enhanced coverage of shifts and the quality of detected shifts, rather than improved speaker attribution. This observation aligns with the findings in Table~\ref{tab:nextspeaker}, which show that diffusion has little effect on oracle-timed who-next accuracy for our system and that its impact is not consistently positive across all methods.

\section{Conclusion}
In this study, we examine full multiparty turn-taking using the VoxConverse dataset with an audio-only, dual-process design. This design suggests candidate end-of-turn decision times in a streaming format and verifies floor transfer (the question of "who speaks next") only at these candidate points. This method improves the detection of \textsc{SHIFT} when compared to a traditional two-stage pipeline baseline. Furthermore, our implementation of label-preserving diffusion augmentation enhances robustness against noisy and overlapping conversational sounds without changing candidate timestamps or turn labels. To our knowledge, this is the first study to use diffusion-generated background mixing for predicting multiparty turn-taking events. Importantly, we provide a controlled dyadic benchmark against VAP by evaluating both systems with identical oracle candidate timestamps and validation-only tuning. This approach helps isolate the effect of verification at decision points.

\bibliographystyle{IEEEtran}
\bibliography{mybib}

\end{document}